# Hyper Heuristic Based on Great Deluge and its variants for Exam timetabling Problem

Ei Shwe Sin[1], Nang Saing Moon Kham[2]

[1]University of Computer Studies, Yangon
*eishwe.ucsy@gmail.com*
[2] Head of Information Science Dept, University of Computer Studies, Yangon
*moonkhamucsy@gmail.com*

## Abstract

Today, University Timetabling problems are occurred annually and they are often hard and time consuming to solve. This paper describes Hyper Heuristics (HH) method based on Great Deluge (GD) and its variants for solving large, highly constrained timetabling problems from different domains. Generally, in hyper heuristic framework, there are two main stages: heuristic selection and move acceptance. This paper emphasizes on the latter stage to develop Hyper Heuristic (HH) framework. The main contribution of this paper is that Great Deluge (GD) and its variants: Flex Deluge(FD), Non-linear(NLGD), Extended Great Deluge(EGD) are used as move acceptance method in HH by combining Reinforcement learning (RL).These HH methods are tested on exam benchmark timetabling problem and best results and comparison analysis are reported.

## 1. Introduction

Today, because of being critical in education sectors, most of the university administrators are trying to get more enrolment of the student and they have to be very careful to increase in student's stratifications. As a result, they are very careful to solve university problem. In fact, it represents the difficult optimization problem. As the difficulty of the problem, their importance in practice and inherent scientific challenge increases, they have been widely investigated across both the operational research and the artificial intelligence community. . It can be classified into exam timetabling problem and course timetabling problem. In this paper, exam timetabling problem (ETP) is used as the test bed for the proposed three HH methods.

The exam timetabling problem is to assign a number of exams to a number of potential time periods or slots by taking into account to satisfy the several constraints. Several approaches have been conducted with various methodologies being applied to attempt to produce better quality exam timetables. There are a lot of researchers and their publications in the literature. For more detailed information about examination timetabling, it can be found in [2, 16, and 18].

There are also varieties of timetabling problem classes on which variety of approaches such as sequential method, cluster methods, constraint-based methods and meta-heuristics are used. Moreover, there are a large number of Meta heuristics for solving an examination timetabling problem. However, these methods have some issues such as parameter tuning and they are not capable of dealing with other different problems. As a result, the current methods being applied to exam timetabling are hyper heuristic (HH). Hyper Heuristic is an emerged search technique for the purpose to raise the generality [7]. Early research works on hyper heuristic focused on the development of advanced strategies for choosing the heuristics to be applied at different points of the search [3]. Likewise, researchers have proposed different acceptance criteria to drive selection of low level heuristics within a hyper heuristic framework. For instance, a Monte Carlo acceptance criterion is used by Ayob and Kendall in [6] while the great deluge acceptance criterion is used by Kendall and Mohamad in [7]. Due to the success of the great deluge and its variants, in this paper, we use them as move acceptance method to find out whether they can support the good quality solutions for the Toronto benchmark exam timetabling problem or not.

The rest of the paper is organized as follows: the Section 2 reviews the previous methods that are related our proposed system while Section 3 describes the exam timetabling problem including constraints. In Section 4, the proposed move acceptance methods are presented. The experiment results and analysis are shown in the Section 5. Finally, the conclusion is shown at the Section 6.

## 2. RELATED WORKS

Normally, a hyper heuristic can conduct with a single point or multi-point search. There are two main stages in a single iteration of a hyper heuristic method .They are heuristic selection and movement acceptance. In this paper, the second stage is emphasized. In general, the movement acceptance can be deterministic or nondeterministic. There are many methods such as Great Deluge (GD), ACO algorithm and simulated annealing methods are used as move acceptance criteria in hyper heuristic because of their very popularities. Therefore, a brief review of GD and its variants is made in this paper.

Bykov Y. proposed the time-predefined great deluge algorithm and Trajectory base search to exam timetabling in 2003 [19] and Edmund K. Burke and Yuri Bykov made an extension of the great deluge algorithm (which they called "Flex-Deluge") where the acceptance of uphill moves depends on a "flexibility" coefficient, for solving exam timetabling problem in 2006 . Good results were presented and they suggested that the flex deluge method is relatively higher effective in the large-scale problems [8].

In 2007, C. Pramodh and V. Ravi also proposed four variants of Modified Great Deluge Algorithm based Auto Associative Neural Network (MGDAAANN) and worked on three different banks data sets [4]. Likewise, Bilgin et al. also reported that a simple random-great deluge based hyper heuristic was the second best after choice function-simulated annealing, considering the average performance of all hyper heuristic over a set of examination timetabling problem[1]. For course timetabling problem, non linear great deluge algorithm (NLGD) was proposed by Landa Silva and Obit. That method produced new best in 4 out of 11 course timetabling problem instances of datasets [14]. In addition , McMullan proposed an extended great deluge algorithm(EGD) for university course timetabling , which allows re-heating similar to simulated annealing, and found new best results for the 5 medium instances. Moreover, in 2009, the EGD algorithm is also investigated and made a comparison with the first winner, Tomas Muller in the 2nd International Timetabling Competition (ITC2007). And it seems that EGD is comparable to existing state of the art techniques, and form previous application to other data sets and a different problem domain (course timetabling)[3].

In 2010, Nabil Nahas , Mustapha Nourelfath and Daoud Ait-Kadi have proposed the Iterated great Deluge heuristic to implement for the dynamic facility layout problem(DFLP) . It consists of two main steps. The objective of the first step is to find a local optimum solution by EGD and the second step is a loop that allows the search process to alternate between diversification and intensification [18]. Likewise, in 2010, Ender Ozcan et al. proposed a hyper heuristic system by using reinforcement learning in heuristic selection and great deluge in move acceptance method and also achieved the comparative results with other HH methods in the literature [9]. By following this idea, we have already proposed to employ the Extended Great Deluge (EGD) as move acceptance method to make a decision whether to accept or reject a resultant solution in RL based HH framework. Therefore, from these well-known literature reviews and experiences, now, we would like to do another contribution by making comparison and more analysis about the GD and its other variants in HH framework.

## 3. DESCRIPTION OF THE EXAM TIMETABLING PROBLEM

The university exam timetabling problem can be defined and described in many ways. The basic way to represent to it is graph model .Mathematical model can also be used. In a more formal way, the timetabling literature defines two types of constraints. Hard Constraints are the constraints that must be satisfied at all times. Soft Constraints are not critical but their satisfaction is beneficial to students and/or the institution. Typically one cannot satisfy all soft constraints thus there is a need for a performance function measuring the degree of satisfaction of these constraints [3]. The primary hard and soft constraints in exam timetabling problem can be found in [2]. Among them, the following table shows the hard and soft constraints are used in this paper.

Table 1. Hard and Soft Constraints

| Constraints | Description |
|---|---|
| HC1 | No exams with common resources (e.g. students) assigned simultaneously. |
| HC2 | Resources of exams need to be sufficient (i.e. size of exams need to be below the room capacity, enough rooms for all of the exams.) |
| HC3 | Each examination must be assigned to a timeslot only for once. |
| HC4 | All the examinations must be scheduled. |
| SC1 | A student should have at least a single timeslot in between his/her examinations in the same day. |

In addition, the problem can be defined with the terms shown in the table 2.

Table 2. Problem Description

| E | the number of n exams: $E_1, E_2, E_3, \ldots, E_n$ |
|---|---|
| S | the number of m students: $S_1, S_2, S_3, \ldots S_m$; |
| T | the number of k timeslots: $T_1, T_2, T_3, \ldots T_k$; |
| $B_{n*k}$ | A binary matrix such that $b_{ik}=1$ when exam $e_i$ is assigned to the timeslot $t \in T$ and $b_{ik}=0$ otherwise. |

| $C=(c_{ij})_{n*n}$ | The conflict matrix; where each element (denoted by $c_{ij}$ where i, j) is the number of students that have to take both exams i and j. This is a symmetrical matrix of size N, where diagonal element $c_{ii}$ equal the number of students who have taken exam i. |
|---|---|

In our system, we use not only the conflict matrix but also the binary matrix to ensure HC1 and HC3. The objective is to schedule all of the exams into time slots, while minimizing the average total cost per student. The following function is used to calculate the average cost per student.

$$\frac{\sum_{i=1}^{n} w_i(|e_i - e_j|)c_{ij}}{s} \quad (1)$$

In equation, $w_i$ is the weight that represents the importance of scheduling exams with common students i timeslots apart , where, w(1)=16, w(2)=8 w(3)=4,w(4)=2 and w(5)=1, i.e. the smaller the distance between periods the higher the weight allocated. Note for n>5, w (n) =0. For example, if a student has two consecutive examinations (i.e. no free time between them) then the weight value of 16 is assigned. If a student has two consecutive exams with a free timeslot in between then a value 8 is assigned and so on. The value of $c_{ij}$ is the number of students common to both examinations. The example of conflict matrix is presented in the following figure 1. In the figure , the value 3 of $c_{11}$ means that there are 3 students who takes the exam E1 and the value 1 of $c_{12}$ means that there is 1 conflict student who will take not only the exam E1 but also exam E2. This conflict matrix is also symmetric.

|    | E1 | E2 | E3 | E4 | E5 | E6 | E7 |
|----|----|----|----|----|----|----|----|
| E1 | 3  | 1  | 2  | 1  | 2  | 1  | 1  |
| E2 | 1  | 3  | 0  | 1  | 2  | 2  | 1  |
| E3 | 2  | 0  | 2  | 1  | 1  | 0  | 1  |
| E4 | 1  | 1  | 1  | 2  | 1  | 0  | 1  |
| E5 | 2  | 2  | 1  | 1  | 3  | 1  | 0  |
| E6 | 1  | 2  | 0  | 0  | 1  | 2  | 1  |
| E7 | 1  | 1  | 1  | 1  | 0  | 1  | 2  |

Figure 1. Example of Conflict Matrix

## 4. PROPOSED GD AND ITS VARIANTS FOR MOVE ACCEPTANCE IN HH METHODS

By using reinforcement learning in the first stage of HH and the variants of GD are used as move acceptance method, the three hyper heuristic systems such as: RL_EGD, RL_NLGD and RL_FD are discussed in this section. Firstly, we present how to produce an initial solution and its representation and the low level heuristic which are used in these proposed systems. And then, the analysis and comparison of these three HH systems are described.

## 4.1. Low Level Heuristic and Initial Solution

To get the final optimized solutions, it also totally depends on the set of heuristics it can be chosen from. Also, due to the performance changes of a number of heuristics over a search space, it is not easy to find a heuristic that always produces the best decisions. They are heuristics that allow movement through a solution space and that require domain knowledge and are problem dependent. Each heuristic creates its own heuristic search space that is part of the solution search space. There are many low level heuristics (LLH) in the literature, for example: mutational heuristic, ruin or recreate heuristic and so on. For low level heuristic module, the following table is presented the low level heuristics used in this paper.

Table 3. Low level Heuristics

| Low level Heuristic | Description |
| --- | --- |
| Largest Enrolment-(LE) | This heuristic takes exams with the largest number of registered students and schedules them first. |
| Swap Timeslot with Kempe Chain (ST-KC) | Swapping a subset of exams in two distinct timeslots making sure that a hard constraint violation does not occur. |
| Reassign Timeslot (RT) | Randomly reassign a sequence of timeslots. |
| Inverting Timeslot (IT) | Inverting timeslots making sure that a hard constraint violation does not occur. |
| Shifting Timeslot (ST) | Left or right shifting a sequence of timeslots. |

The first heuristic is one of the graph colourings heuristic and which is used to create a feasible initial solution in this paper. Generally, it is important to have an easy and quick way of generating an initial solution. Note that it is not necessary though that initial solution should be completely feasible. However, it is preferred to be as feasible as possible because the quality of initial solution would affect the final solution. Therefore, in this paper, completely feasible solution is produced by using (LE) heuristic. The examination with the largest student enrolment is selected and kept in the vector. And then another exams which are not conflict with the exam in the vector, are added to that vector. This process is repeated until the required capacity for exams in the vector is less than the total capacity and unscheduled exam list is empty. Finally, this exam vector which satisfied all hard constraints is assigned to the timeslot sequentially. For the solution representation; there are two forms of assignment such as:

- Exam-Timeslot assignment
- Exam- Classroom assignment.

In this paper, the first one is used to represent of the solution. By taking the advantages of HH method, to produce the feasible solution of all data instances is the aim of this paper. Moreover, one of the contributions of this paper is that the average timeslot per exam is also used to be balance in exam-timeslot assignment. Otherwise, many exams are assigned to a timeslot whereas few exams are assigned in another timeslot.

The rest four low level heuristics from the table 3 are used in the reinforcement learning process to improve the solution to get the optimal solution which meets the objective. These heuristics are also called the slot move in [2]. The following figure shows the swapping a set of assigned exams in two distinct timeslot as an example. In figure, the exams e2, e3, e5 are assigned timeslot 5 whereas the exams e4 and e1 are assigned in timeslot 2. After the heuristic, ST-KC has been applied; the set of exams assigned in each timeslot will be changed.

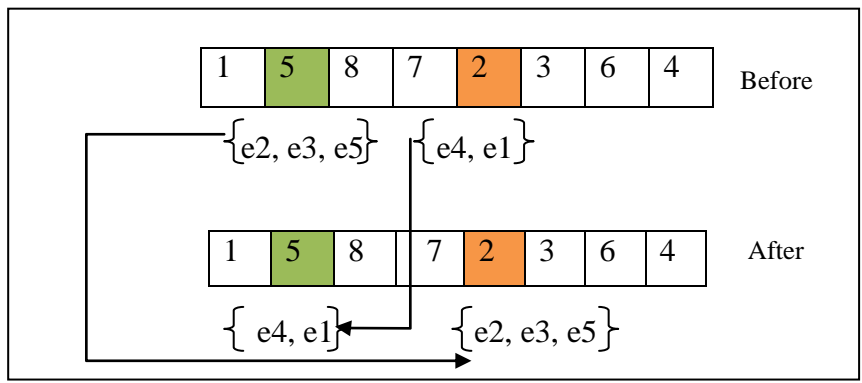

Figure 2. Example of Swap Timeslot with Kempe Chain (ST-KC)

### 4.2. Reinforcement Learning

For the heuristic selection process in hyper heuristics, machine learning techniques are vital to make the right choices. Learning can be achieved in an offline or online manner. An online learning hyper heuristic learns through the feedback obtained during the search process while solving a given problem. In addition, it is better than offline method. Most of the existing online learning hyper heuristic incorporates reinforcement learning (RL). It is a sub-field of machine learning, represents an important direction for research in Artificial Intelligence [9]. RL is a framework for learning an optimal policy of a task from trials. It requires less a priori knowledge. Furthermore, it is successfully applied to scheduling, control, game theory and so on. However, the quality of solution obtained by using RL is not satisfactory in many times. However, EGD can control to make a decision whether it is accepts or rejects, after the chosen heuristic has been applied to initial solution. In this paper, not only a simple P: 1-N: 1(Additive adaption-Negative adaption) strategy to increase and decrease the utility value and but also maximum utility method are used for RL. For the parameter setting for lower bound and upper bound of utility values in RL, we follow the reference [9] and the values can be seen at the Section 5.

### 4.3. Great Deluge (GD) and its variant

The Great Deluge algorithm (GD) is a genetic algorithm applied to optimization problems. It is similar in many ways to the hill-climbing and simulated annealing algorithms. In GD, the water level is set to a value higher than the expected penalty of the best solution at the start of the search. Then the water level is decreased in a linear fashion during the search until it reaches a value of zero [19]. It is a well known acceptance method proposed by (Dueck; 1993, Burke et al., 2003). There are many variants form of GD in the literature such as NLGD, EGD and FD. They are discussed in the next section in detail.

#### 4.3.1. Extended Great Deluge

In fact, the concept of EGD algorithm is quite similar with the hyper heuristic method. As far as the author knows, it has been considered first time as move acceptance for hyper heuristic. It has advantages to require the tuning of a few input parameters that can represent the search time. It can provide a wider test with the hidden data sets for consistency in the approaches. Moreover, it can be interested to run all techniques at some future point with further hidden data sets. It is also proved to be both robust and general. Because of these advantages, it has been successfully applied to many optimization problems such as buffer allocation problem, redundancy allocation problem and so on.

Therefore, in this paper, it is investigated to make further improvement in hyper heuristic or not. The standard GDA has been extended by adding reheat mechanism, step 13 in figure 1, similar to that employed with simulated annealing in timetabling. The aim of this approach is to both improve the speed at which an optimal solution can be found and at the same time utilize the benefits of this technique in avoiding the trap of local optima. In addition, the Great Deluge generally can cause the continuous lack of improvement, which means the final solution is same with the initial after the complete execution, which can lead to RL to select only one heuristic repeatedly. Rather than terminating, the extended approach employs reheating in order to relax the boundary condition to allow worse moves to be applied to the current solution. Cooling continues and the boundary is reduced at a rate according to the remaining length of the run. The initial decay rate of the EGD is used to show how fast the boundary is reduced and ultimately the condition for accepting worse moves is narrowed. In this paper, we use the half-life decay rate, is the amount of time it takes for half of the amount of substance to decay to attempt to reach the optimal solution. The wait parameter is used to invoke the reheat mechanism. It can be specified in terms of percentage or number of total moves in the process [3].

### 4.3.2. Flex-Deluge

An extension of the Great Deluge algorithm is "Flex-Deluge", where the acceptance of uphill moves depends on a "flexibility" coefficient kf (0<= kf<=1). The acceptance rules are outlined in Expression (2):

$$P'=P + kf (B - P) \quad \text{when } P < B \quad P'=P \quad \text{when } P \geq B. \quad (2)$$

by varying kf, it is possible to obtain an algorithm with characteristics of both the original Great Deluge (kf = 1) and greedy Hill-Climbing (kf = 0). This method enables the search procedure to develop with an adaptive level of strictness of acceptance for each particular move [8]. Thus, in this paper, the flexibility coefficient value for the all data instances can be seen in the Section 5.

### 4.3.3. Non-Linear Great Deluge

Another extension of GD is non-linear great deluge algorithm (NLGD) in which the acceptance criterion refers to accepting improving and non-improving low-level heuristics depending of the performance of the heuristic and the current water level B. Improving heuristics are always accepted while non-improving ones are accepted only if the detriment in quality is less than or equal B. The initial water level is usually set to the quality of the initial solution and then decreased by a non-linear function proposed in [17] as follows:

$$B = B \times (\exp^{-\delta} (\text{rnd } [\min, \max])) + \beta \quad (3)$$

The various parameters in Eq. (3) control the speed and the shape of the water level decay rate. Parameter β influences the shape of the decay rate and it represents the minimum expected penalty corresponding to the best solution. The role of parameters min and max is to control the speed of the decay rate. Therefore, for higher values of min and max, the water level decreases more rapidly and hence, improvements to the solution quality are also achieved faster [14]. As far as the author knows, it has been employed only for the university course timetabling problem [5, 13, 14 and 16]. So, this paper is first used and tested for exam timetabling problem. The parameter values are being set not to be specified for all data instances and these are shown in table 4.

## 4.4. Three HH Methods for Exam Timetabling Problem

The first HH method, RL-EGD has already been proposed and published as our previous job [10]. It is capable of producing feasible solutions for all problem instances and comparable results in the literature. Now other two HH methods such as RL-FD and RL-NLGD for exam timetabling problem are implemented and compared with it. The main differences of these three HH methods can be seen at the step 10 of the figures 3, 4 and 5.

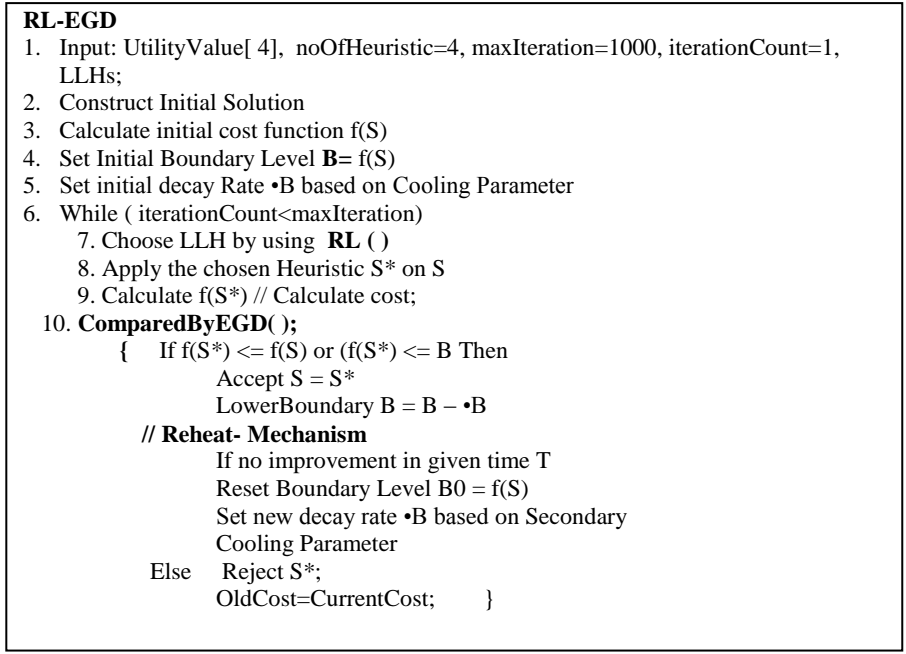

Figure 3. RL-EGD HH framework

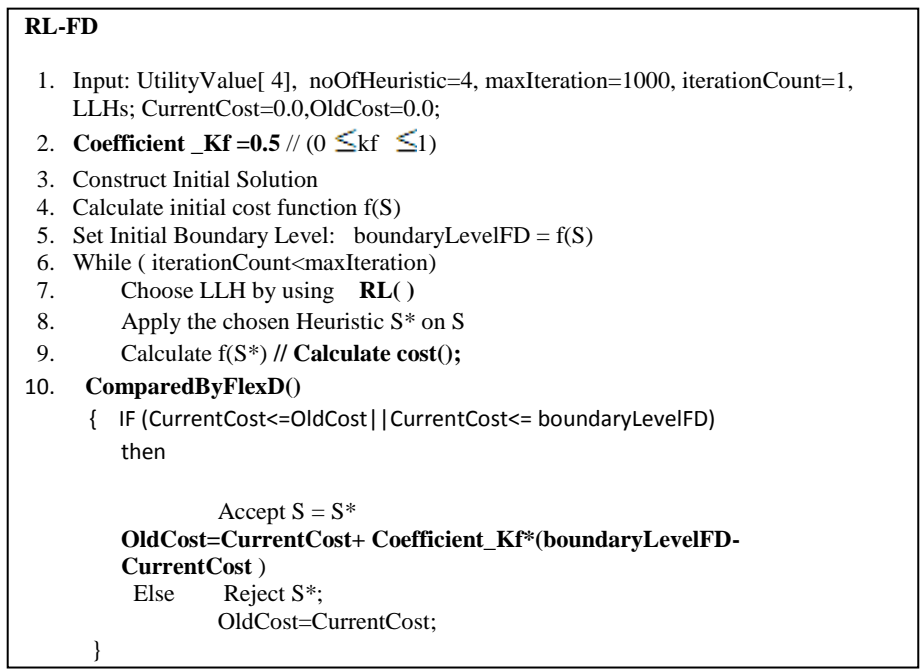

Figure 4. RL-FD HH framework

```
RL-NLGD

1.  Input: utilityValue[4],noOfHeuristic=4,maxIteration=1000, iterationCount=1,
    LLHs; CurrentCost, OldCost, β =0.0,Bmin =100000,  Bmax =300000, δ =5*10⁻¹⁰
2.  Construct Initial Solution
3.  Calculate initial cost function f(S)
4.  Set Initial Boundary Level boundaryLevelNLGD=OldCost;
5.  Set initial decay Rate
6.  While ( iterationCount<maxIteration)
     7.   {  Choose LLH by using   RL ( )
     8.   Apply the chosen Heuristic S* on S
     9.   Calculate f(S*)
     10.  ComparedByNLGD()
          {  If(CurrentCost<=OldCost||CurrentCost<= boundaryLevelNLGD) then

               Accept S = S*
               B = B*(exp−δ (rnd[min,max]))+β

            Else   Reject S*;
                 OldCost=CurrentCost;

          }
```

Figure 5.  RL-NLGD Framework

## 5. RESULTS AND DISCUSSION

Our experimental analysis is making on the computer Pentium IV, Dell with the RAM 2GB .We tested the three HH methods on 31 instances. These are 13 instances proposed by Carter et al. and 18 instances created by instance generator. All of these data instances and random instance generator can be available from the link: "http://www.asap.cs.nott.ac.uk/resources/data.shtml.

Each data instance is executed 10 times as shown in the following table. Each run is performed starting from the same initial configuration. The HH methods are implemented in Java and the parameters and environmental settings are as shown in the following table.  In the parameter setting, Coefficient_Kf is used in the RL-FD while β, $B_{min}$ , $B_{max}$, δ are being used in RL-NLGD. For the values of these parameters, we follow the reference [16] because that paper provided the better results.

Table 4. Parameter Setting Values

| *Parameters* | *Value* |
|---|---|
| Number of runs | 10 per each data instances |
| Number of Iterations | Maximum Iteration=1000 |
| Coefficient _Kf | 0.5 |
| β | 0.0 |
| $B_{min}$ | 100000 |
| $B_{max}$ | 300000 |
| $\delta$ | $5*10^{-10}$ |
| Wait Value for reheat mechanism | 25% |
| Lower bound | 0.0 |
| Upper Bound | 40.0 |
| Utility Value for each low level heuristic | 0.75*Upper Bound |

To make a comparison, the **Lowest Best Cost** is used as the performance criterion for all experiments. According to the objective function, the lowest the cost, the better the timetable is. The results of lowest best cost for each HH methods are presented in the following graphs.

The figure 6 shows the comparison of the Lowest Best Cost on small 9 data instances. At the same time, the figure 7 shows the comparison on large 9 data instances. Although there is no significant difference in the cost values of three HH methods in each dataset, it is observed that RL-EGD can provide the lowest best cost in small five data sets whereas the other two HH methods, RL-FD and RL-NLGD can produce the lowest cost in two datasets. Likewise, it is observed that RL-EGD can also provide the lowest best cost in large five data sets. The figure 8 shows the comparison for 5 data instances from the Carter's Datasets. We choose these data instances by randomly. Here, because of the reheat mechanism of the RL-EGD, it can also produce much lower costs in three instances than those by other two HH methods. In these figure, the symbol star shows the data instance got the lowest best cost by RL-EGD HH approach.

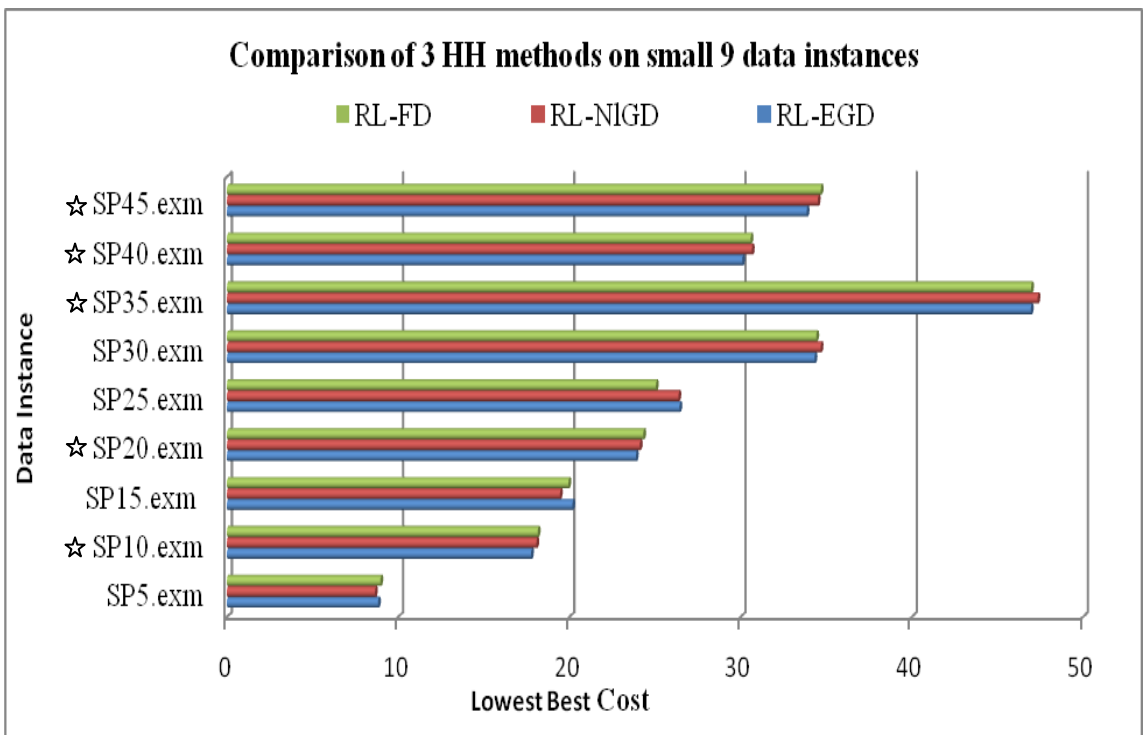

Figure 6. Comparison of Three HH methods on Small 9 Data Instances generated by Random Generator

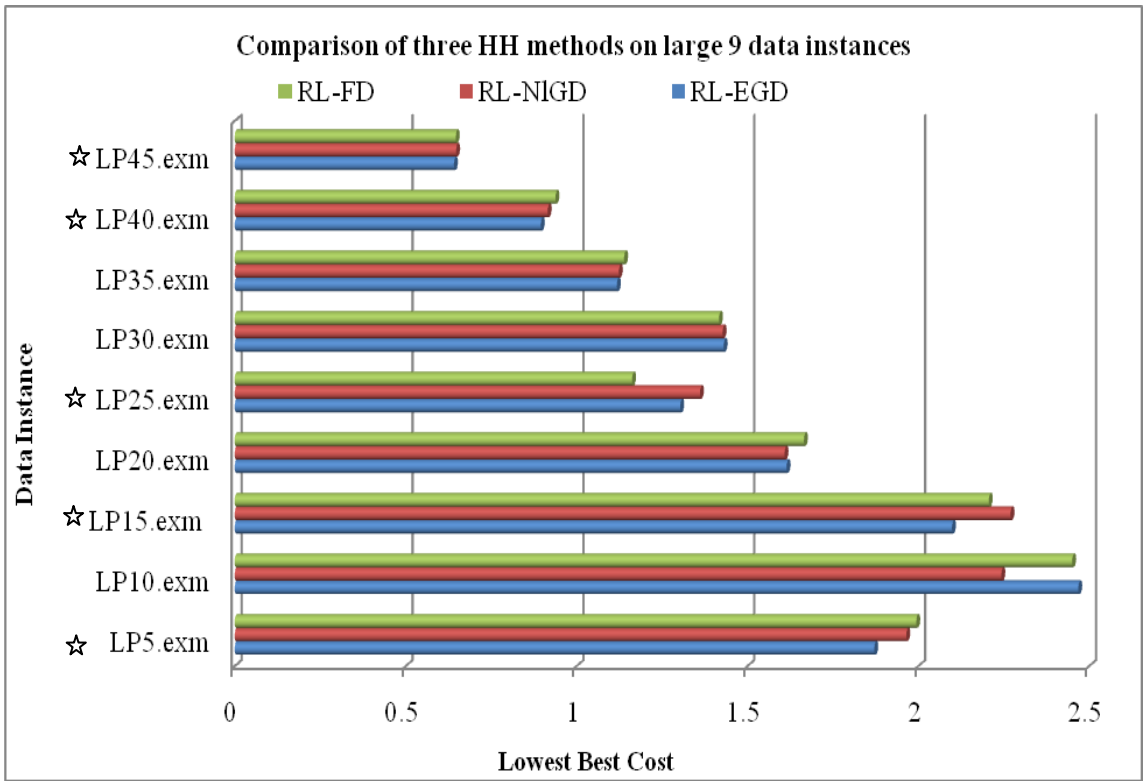

Figure 7. Comparison of Three HH methods on 9 large data instances generated by Random Generator

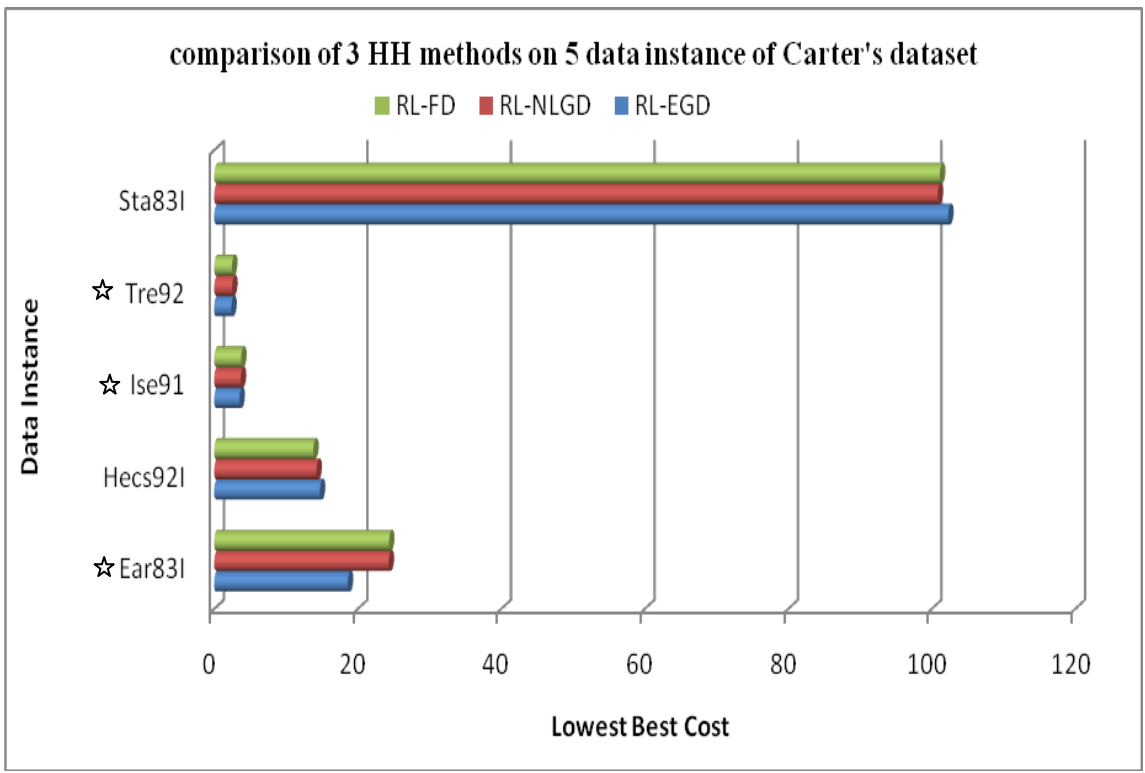

Figure 8. Comparison of three HH methods on 5 data instances from Carter's Dataset

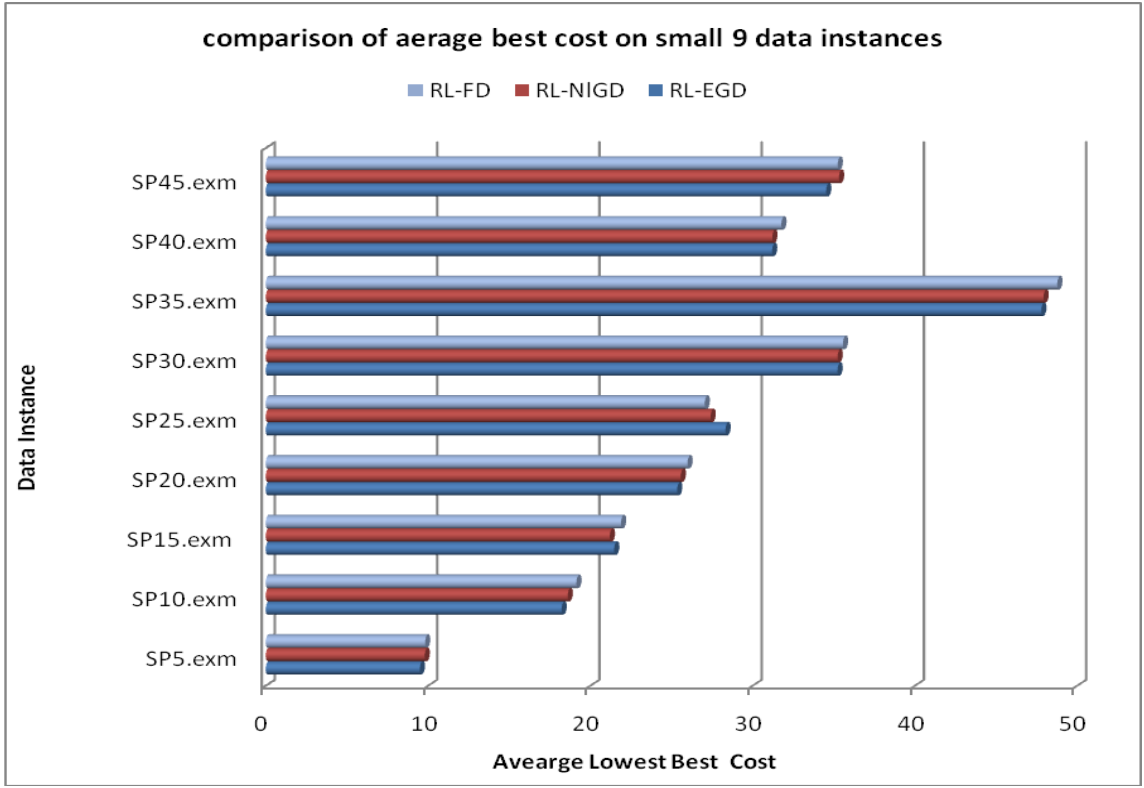

Figure 9. Comparison of Average Lowest Best Cost

For another comparison, the figure 9 represents the comparison of average lowest cost of all small data instances. Among three HH methods, it can be seen that RL-EGD methods has achieved the lowest average best cost in most of the data instances except the SP25.exm and SP15.exm. In figures, X axis shows the data instances names while Y axis is being representing the lowest cost of the best solution.

## 6. CONCLUSION

Hyper heuristics have been starting to prove themselves as fast and effective methods for solving complex real world optimization problems. Therefore, the RL-EGD HH method is also proposed as our first job and also achieves the better solutions for the problem domain. Now, as the next step, we have made a comparison it with two other HH methods. From the experiments, it can be concluded that the method RL-EGD can provide the lowest best cost for most of the data instances and compare with the other methods in the literature. To be able to schedule invigilators such as professors, doctors and teaching assistant to proctor the scheduled exams, as the future work, another enhancement would be to integrate this system with the invigilation timetabling system.

### Acknowledgements

I would like to express my gratitude to my rector, Dr. Ni Lar Thein, and all of my honourable teachers who check my grammar errors and provide valuable suggestions and guidance.

## REFERENCES

[1] B. Bilgin, E. Ozcan, & Korkmaz E. E.(2007), "An experimental study on hyper heuristic and exam timetabling", In(PATAT'06)(LNCS 67,pp.394-412).

[2] B. Bullnheimer, "An Examination Scheduling Model to Maximize Students' Study Time", Practice And Theory of Automated Timetabling (PATAT's 97), August 1997, Toronto

[3] B. McCollum, P.J. McMullan, A. J. Parkes, E.K. Burke, S. Abdullah, "An Extended Great Deluge Approach to the Examination Timetabling Problem ", MISTA,2009.

[4] C. Pramodh and V. Ravi, " Modified Great Deluge Algorithm based Auto Associated Neural Network for Bankruptcy Prediction in Banks", International Journal of Computer Intelligence Reserach, ISSN 097-1873 Vol.3, No.4(2007), pp. 363-370

[5] Dario Landa-Silva and Joe Henry Obit, "Evolutionary Non-linear Great Deluge for University Course Timetabling"

[6] E. Burke, G. Kendall, E. Soubeiga. "A tabu-search hyper heuristic for timetabling and roistering", Journal of Heuristic, 9:451-470, 2003.

[7] E. K. Burke, Kendall Graham and R. Ozcan," Hyper Heuristics: A Survey of the State of the Art". Journal of the Operational Research Society, Palgrave Macmillan.

[8] E. K. Burke, Y. Bykov, "Solving Exam Timetabling Problems with Flex-Deluge Algorithm", PATAT 2006, pp. 370-372. ISBN 80-210-3726-1.

[9] E. Ozcan, M. Misir, G. Ochoa, E. K. Burke , "A reinforcement learning- great deluge hyper heuristic for Examination Timetabling" , In: Journal of Applied Met heuristic Computing, 1(1), 39-59, January-March 2010.

[10] ES Sin, "Reinforcement Learning with EGD based Hyper Heuristic System for Exam Timetabling Problem", In: Proceedings of the IEEE-CCIS 2011.

[11] E. Soubeiga. "Development and application of hyper heuristic to personnel scheduling." PhD thesis, School of Computer Science, University on Nottingham, UK, 2003.

[12] Gabriela Serban , " A new reinforcement learning algorithm", STUDIA UNIV. BABES-BOLYAI, INFOMATICA, Volume XLVIII, Nubmer 1,2003.

[13] H. J. Obit, D. Landa-Silva,D. Quelhadj , M. Sevaux, "Non-Linear Great Deluge with learning Mechanism for Solving the Course Timetabling Problem", MIC-2009: The VIII Meta-heuristic International Conference.

[14] Joe Henry Obit et al.,"Non Linear Great Deluge with Reinforcement Learning for University Course Timetabling".

[15] K. Leslie and Michael L. Littman, "Reinforcement learning: A Survey", Journal of Artificial Intelligence Research 4 (1996) 237-285.

[16] Landa-Silva, Obit, H.H: "Great Deluge with Nonlinear Decay Rate for Solving Course Timetabling Problem", In: Proceedings of the 2008 IEEE Conference on Intelligent System (IS 2008), pp. 8.11-8.18. IEEE Press, Los Almitos (2008)

[17]   M. Carter and G. Laporte:" Recent Developments in Practical Exam Timetabling". In: Burke E.K and Ross P. (eds.): (PATAT95) Lectures Notes in Computer Science 1153, 3-21, 1996.

[18]   Nabil Nahas, Mustapha Nourelfath and Daoud Ait-Kadi," Iterated Great Deluge for the Dynamic Facility Layout Problem", in May, 2010.

[19]   Y. Bykov,"Time-Predefined and Trajectory-Based Search: Single and Multiobjective Approaches to Exam Timetabling", 2003.

**Authors**

Ei Shwe, Sin

Ph.D(IT) Research Student

University of Computer Studies, Yangon

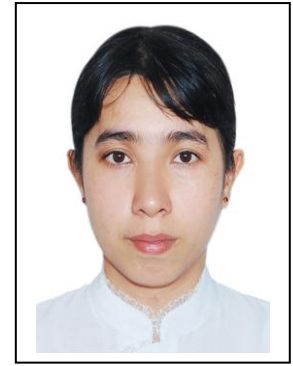